# Multimodal Training to Unimodal Deployment: Leveraging Unstructured Data During Training to Optimize Structured Data Only Deployment

Zigui Wang, BS.[1], Minghui Sun, MB.[1], Jiang Shu, MB.[1], Matthew M. Engelhard MD, PhD[1], Lauren Franz, MBBCh[3,4], Benjamin A. Goldstein, Ph.D[1,2]

[1]Department of Biostatistics and Bioinformatics, Duke University School of Medicine, Durham, NC, USA;
[2]Department of Pediatrics, Duke University School of Medicine, Durham, NC, USA;
[3]Duke Center for Autism and Brain Development, Duke University School of Medicine, Durham, NC, USA;
[4]Department of Psychiatry and Behavioral Sciences, Durham, NC, USA;

## Abstract

*Unstructured Electronic Health Record (EHR) data—such as clinical notes—contain clinical contextual observations that are not directly reflected in structured data fields. This additional information can substantially improve model learning. However, due to their unstructured nature, these data are often unavailable or impractical to use when deploying a model. We introduce a multimodal learning framework that leverages unstructured EHR data during training while producing a model that can be deployed using only structured EHR data. Using a cohort of 3,466 children evaluated for late talking, we generated note embeddings with BioClinicalBERT and encoded structured embeddings from demographics and medical codes. A note-based teacher model and a structured-only student model were jointly trained using contrastive learning and contrastive knowledge distillation loss, producing a strong classifier (AUROC = 0.985). Our proposed model reached AUROC of 0.705—outperforming the structured-only baseline of 0.656. These results demonstrate that incorporating unstructured data during training enhances the model's capacity to identify task-relevant information within structured EHR data, enabling a deployable structured-only phenotype model.*

## Introduction

Electronic Health Records (EHRs) are a rich source of clinical information and are increasingly leveraged for clinical modeling and decision support[1,2]. EHR databases contain diverse data modalities, which can be categorized as (a) structured or tabular data such as demographics, diagnoses, procedures, and laboratory values; and (b) unstructured data such as clinical notes, images and genetic sequencing data[3]. In general, structured data are analytically more straightforward to work with because they are typically easier to access, consist of well-defined organized variables, and lead themselves readily to standard statistical and machine learning methods [4]. This helps explain why most EHR based analytic tools rely heavily on these modalities[5,6]. However, there is often rich information contained within the unstructured data which can add important contextual information to clinical analyses.

Newer machine learning based methods, like natural language processing and representation learning, have substantially improved the ability to process unstructured data by extracting informative embeddings from raw text, images, and signals. As a result, incorporating unstructured data into analytic tasks is more feasible[6,7]. However, while incorporating such data into retrospective analyses is becoming more standard, there are still challenges to using them for both retrospective and prospective analytics. For example, clinical notes, which contain identifiable information, may not be easily shared across institutions; imaging data may not be available at the time an analytic assessment is performed; and genomic sequencing data are often missing because they are collected for only a subset of patients. As such it is important to develop analytic methods that learn from and incorporate insights from unstructured data without necessarily having those data elements available at the time of assessment.

A typical analytic task is scanning across a patient's medical record to determine what conditions they have. For retrospective research, this is referred to as "computable phenotyping." Computable phenotyping has been well studied, with both simple (i.e., Boolean based rules) and complex (i.e., probabilistic models) algorithms being used [8–11]. Among the various EHR modalities, clinical notes are particularly useful for aiding diagnosis, as they include a more complete set of provider impressions of the patient. Unlike structured EHR data that primarily record coded diagnoses, procedures, or laboratory values, clinical notes capture physicians' reasoning, contextual details, and patient-specific nuances. These texts often document symptom trajectories, social and behavioral factors, and clinical impressions that may be absent from structured data[12,13]. To leverage the information within notes, natural language processing methods including large language models (LLMs) have been applied to transform clinical narratives into predictive representations[14,15]. Incorporating clinical notes alongside structured data has been shown to improve sensitivity and specificity in computable phenotype identification, especially for complex or under-documented conditions[16].

Given the performance advantage of multi-modal EHR data, there has been a growing interest in multimodal learning approaches for computable phenotyping[3,17]. These methods typically employ separate encoders for each modality and then integrate the resulting representations through fusion strategies such as attention mechanisms or contrastive learning. For instance, structured EHR data are often modeled using traditional multilayer perceptrons or recurrent neural networks, whereas transformer-based language models such as ClinicalBERT[18], BioBERT[19], and MedBERT[20] are applied to process clinical narratives. However, most multimodal approaches assume that clinical data modalities are complete during both training and inference stages[21]. Nevertheless, as discussed, this assumption does not always hold in clinical practice[22,23]. EHR data are often collected by different parties or maintained in separate systems, leading to inconsistent availability across contexts. For instance, imaging studies performed at an outside facility may not be imported into the local EHR, even when the corresponding radiology report is available.

Some work has begun to address this challenge. For example, cross-modal generation and reconstruction methods, such as generative adversarial networks (GANs) or variational autoencoders (VAEs), aim to reconstruct the missing modality from the available ones[24]. Shared latent–space approaches, such as CLIP[25], learn modality-specific encoders that map inputs (e.g., text and images) into a joint embedding space, allowing representations from different modalities to be directly compared[26]. Finally, knowledge transfer frameworks train a "teacher" on the full multimodal data and distill its learned information to a "student" model that can operate with fewer modalities[3,27].

Most of the work in this space has been conducted in the computer vision-language setting, whereas less has focused on clinical diagnostics and point of care assessment. One recent exception is CLOPPS[22], which primarily focuses on longitudinal data and treats all modalities as equally informative and interchangeable, aiming to learn cross-modal alignment. In contrast, our study addresses a more asymmetric and clinically realistic scenario: structured EHR data serve as the primary modality for deployment, while unstructured clinical notes are used only during training. This design reflects the real-world availability of EHR modalities, where structured data are consistently accessible but unstructured text often is not. Furthermore, beyond contrastive alignment, we also incorporate knowledge distillation framework to explicitly transfer task-relevant predictive information from the note-based teacher model to the structured-data student model, enabling improved performance at inference.

The goal of this paper is to further motivate and develop a teacher-student multi-modal approach, named C-CKD (Contrastive and Contrastive Knowledge Distillation), that is able to learn from complex modalities during training (the teacher) but function effectively with simpler modalities only (the student). We apply our framework to the task of identifying children with late language emergence, commonly referred to as "late-talking" (LT). LT is relatively common, affecting 10-20% of young children[28]. We have recently shown that while LT is often indicated in clinical notes, there is often under-coding of the condition, via ICD-10 codes. As such, simpler computable phenotypes that rely solely on ICD-10 would be insufficient for identifying LT children, either retrospectively or prospectively[29]. As such approaches that can leverage information from clinical notes, but do not require direct access to the notes would be beneficial.

## Method

***Data Source:*** We extracted data from the Duke University Health System (DUHS) EHR system. DUHS consists of three hospitals and over 150 clinics and has been using an integrated Epic EHR platform since 2013. Clinical data are organized into a research-ready datamart based on the PCORnet Common Data Model[30]. Approximately 85% of children in Durham County, which is a racially and socioeconomically diverse county, receive primary care services through DUHS[31].

***Source Cohort:*** Following previous work[29], we constructed a cohort of LT children and matched controls. We required eligible cases to meet the following criteria: (1) born at DUHS on or after January 1, 2016; (2) have at least two encounters with ICD-10 codes for expressive language delays associated with LT  ('F80.0', 'F80.1', 'F80.2', 'F80.4', 'F80.8', 'F80.81', 'F80.89', 'F80.9', 'R47.9', 'R47.89', 'R62.0', 'R47.82', 'H93.25'); (3) receive the first LT diagnosis during a well-child visit between 12 and 36 months of age; and (4) have a clinical note documented at the diagnostic well-child visit. For each case encounter, a control well-child visit was identified from a child who never received an LT diagnosis, matched on sex, payer type, ethnicity, and age. After applying these criteria, a total of 3,466 patients were included in the study.

*Features Used:* For the structured EHR data, we used demographic features, including sex, age, race & ethnicity, primary language, payer type, diagnosis codes, and procedure codes. Clinical notes consisted of provider notes from the index well child visit, defined as the patient's first documented well-child encounter within the study period.

*Processing Note:* Following the methodology from our previous study[29], we utilized a matched case-control design with an 80/20 training-test split. Clinical notes were processed using BioClinicalBERT (BCB)[18], a transformer-based language model pre-trained on biomedical and clinical corpora, to generate dense vector embeddings. To address the limitation of input sequence length, we employed a Multiple Instance Learning (MIL) framework. In this approach, lengthy clinical notes were segmented into overlapping chunks of 512 tokens. During training, the model processed all chunks for a given encounter, but the loss was calculated and parameters updated based solely on the chunk with the maximum logit (i.e., the most suspicious segment). This allowed the model to learn from the specific portion of the text containing the relevant signal while ignoring irrelevant sections. Additional technical details and model architecture can be found in [29].

**Contrastive Learning**
Contrastive learning seeks to learn representations by aligning embeddings of paired observations (e.g., structured data and notes from the same patient) while separating embeddings from unpaired observations. In this study, we have used the InfoNCE formulation [26]. We consider a batch of N paired samples $\{(x_i, y_i)\}_{i=1}^{N}$, where $x_i$ and $y_i$ denote structured EHR data and note data for patient $i$. If $x_i$ and $y_i$ come from the same patient, we consider it a positive pair, and if not, we consider it a negative pair. We define encoders $f()$ and $g()$ that map the inputs data to normalized embeddings $z_i^x = f(x_i)/\|f(x_i)\|$ and $z_i^y = g(y_i)/\|g(y_i)\|$. The similarity score between two embeddings is defined as $s_{ij} = \exp(\langle z_i^x, z_i^y \rangle / \tau)$, where $\tau > 0$ is the temperature parameter and $\langle z_i^x, z_i^y \rangle$ denotes the inner product between two embedding vectors. The InfoNCE loss can be written as

$$L_{Contrastive} = -\frac{1}{N}\sum_{i=1}^{N} \log \frac{s_{ii}}{\sum_{j=1}^{N} s_{ij}}$$

This formulation encourages each $x_i$ to be closely aligned with its paired $y_i$ while treating all other $y_j (j \neq i)$ in the batch as negatives. A symmetric variant additionally applies the loss in both directions, yielding stronger alignment.

**Contrastive Knowledge Distillation**
Based on our previous study [29], note data achieved excellent predictive performance for late talking prediction (AUROC = 0.99). However, in practice, clinical notes may not be consistently available for new patients. To address this limitation, we employed a knowledge distillation framework to leverage the note data during training while ultimately relying only on structured EHR data at inference. Knowledge distillation transfers predictive knowledge from a high-performing teacher model to a student model [27,32]. In our setting, the teacher model is a multilayer perceptron (MLP) trained on note embeddings, while the student model is an MLP trained solely on structured EHR data.

To enhance knowledge transfer beyond direct logit alignment, we incorporated a contrastive knowledge distillation (CKD) loss function [33]. Specifically, the student and teacher logits were projected into a shared embedding space, and for each patient $i$, the aligned teacher–student pair $(t_i, s_i)$ was treated as a positive pair, while logits from different patients $(s_i, s_j; j \neq i)$ served as negative pairs. The CKD loss encourages the student to align closely with its teacher counterpart while maintaining discriminability across patients. The positive similarity term is defined as the negative Kullback-Leibler (KL) divergence [34] between the teacher and student predicted probabilities of the positive class, where $p(t_i)$ and $p(s_i)$ denote the teacher and student probabilities. The KL divergence is

$$KL(p(t_i)|p(s_i)) = p(t_i)\log\frac{p(t_i)}{p(s_i)} + (1 - p(t_i))\log\frac{1 - p(t_i)}{1 - p(s_i)}$$

Negative-pair similarities are derived from the negative L2 distances between the student logits, which encourages the representations of different patients to remain well separated. The CKD loss is therefore formulated as:

$$L_{CKD} = -\frac{1}{N}\sum_{i=1}^{N} \log \frac{\exp(-KL(p(t_i)||p(s_i))/\tau)}{\exp(-KL(p(t_i)||p(s_i))/\tau) + \sum_{j\neq i}^{N} \exp(-\|s_i - s_j\|_2/\tau)}$$

Where $\tau$ denotes the temperature parameter controlling similarity sharpness.

**Model Architecture**

The overall methodological approach, named C-CKD (Contrastive and Contrastive Knowledge Distillation), is illustrated in Figure 1. We presume that we have two data modalities to train a model and one of those modalities is not available during the model deployment phase. When deploying the model in clinical practice, we rely on the structured data modality, which is generally available for most of hospitals and medical institutions [23,35]. In brief, we take two data modalities (structured EHR and clinical note) as input. For structured EHR, we concatenate the structured demographics (numeric fields) and patient-level medical code (diagnosis code and procedure code) embeddings: codes per patient are ordered by age, mapped with a pre-trained Word2Vec model that is fine-tuned on the cohort, and then aggregated using mean pooling. Then the structured EHR vector is passed to a model composed of a MLP encoder, followed by a linear classifier. Distillation targets come from a fixed MLP trained on note embeddings; we load its precomputed logits. The training objective combines task supervision and contrastive learning or knowledge distillation:

$$L_{C-CKD} = \lambda_{CKD} \times L_{CKD} + \lambda_{Contrastive} \times L_{Contrastive}$$

Where $\lambda_{CKD}, \lambda_{Contrastive}$ are the weighting parameter for loss components.

**Experiment**

For the experiment, we consider two baseline models, namely C-CKD Teacher and EHR Data Only models. For the C-CKD Teacher, both structured EHR and note data are available in the training and deployment, which represents the ideal scenario in which no data are missing. For the EHR only model, we use only structured EHR in the training and deployment phase, which represents the worst-case scenario in which note data are not available during training or deployment. Our loss function used for training is comprised of two components: the contrastive alignment loss ($L_{Contrastive}$) and the contrastive knowledge distillation loss ($L_{CKD}$). To better understand the contribution of each objective, we conducted an ablation study that isolates their individual effects. Specifically, in addition to the C-CKD model—which jointly optimizes both losses to align cross-modal representations (contrastive loss) and transfer task-specific information from the teacher to the student (CKD loss) —we introduced two ablation variants: a Contrastive model and a CKD model. The Contrastive model was trained using the contrastive alignment loss combined with a binary cross-entropy (BCE) loss. Since the contrastive objective primarily encourages embedding-level alignment rather than direct predictive optimization, the BCE term ensures the model remains supervised for the downstream classification task. This variant allows us to assess whether cross-modal embedding similarity alone can improve the student model's performance without guidance from the teacher model. Conversely, the CKD model omits the contrastive alignment term and instead relies solely on the knowledge distillation objective, transferring task-specific decision boundaries from the teacher model to the student. By comparing the performance across these three variants—the C-CKD model, the Contrastive model, and the CKD model—we can quantify the distinct and complementary contributions of representation alignment and knowledge transfer to overall model performance.

The performance of all models was evaluated in terms of outcome classification accuracy, as quantified using the area under the receiver operating characteristic curve (AUROC) [36]. For each AUROC result, 95% confidence intervals were estimated using 1,000 bootstrap resamples of the test dataset. In addition to the AUROC, we further evaluated each model's performance across varying decision thresholds ($\alpha$ values) using a set of decision rule diagnostic metrics, including coverage, sensitivity, specificity, positive predictive value (PPV), negative predictive value (NPV), and overall accuracy. Coverage quantifies the proportion of samples for which the model provides a prediction under a given threshold, whereas sensitivity and specificity measure the model's ability to correctly identify positive and negative cases, respectively. PPV and NPV assess the reliability of the predicted outcomes, and accuracy summarizes the overall correctness of classification decisions. $\alpha$ was varied across five probability thresholds of 0.1, 0.2, 0.3, 0.4, and 0.5, respectively.

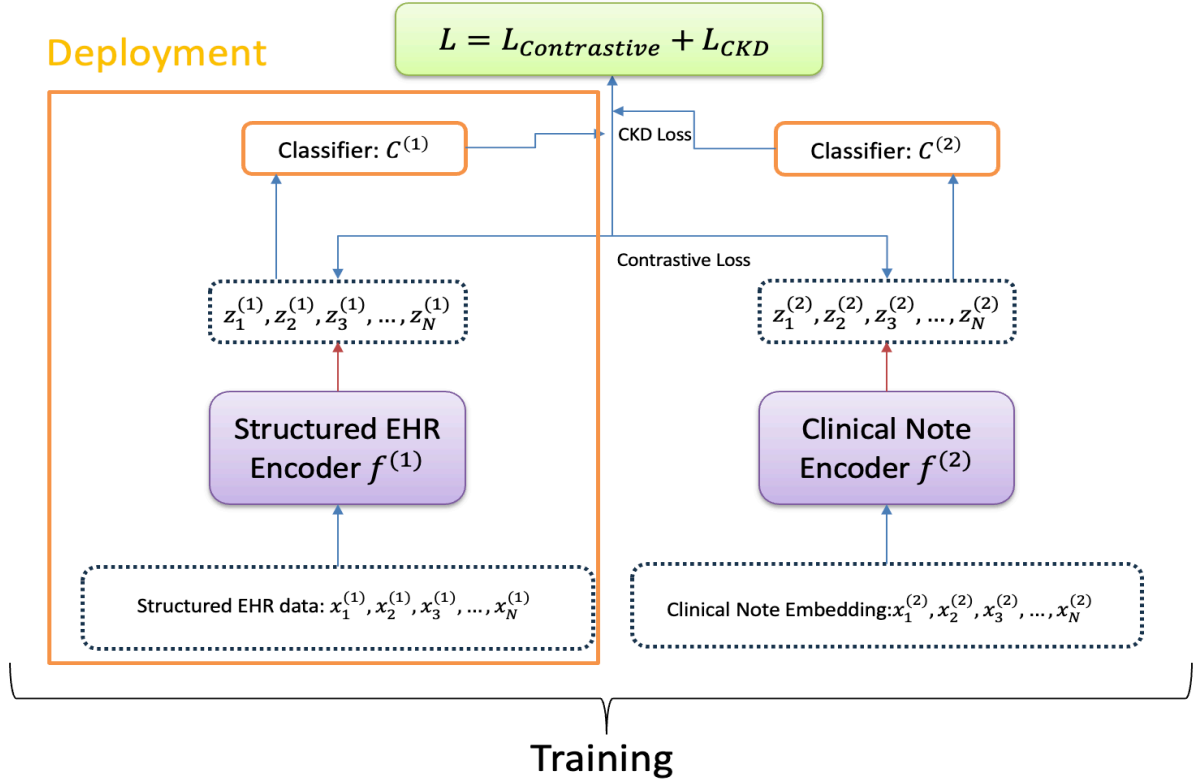

**Figure 1: Diagram illustrating our knowledge distillation model architecture.**

# Results

Table 1 summarizes the demographic characteristics of both the full cohort (n=3,466) and the held-out deployment (test) sample (n=694). The AUROC and 95% confidence interval for C-CKD teacher, C-CKD student, EHR Data Only, CKD model, and Contrastive model are shown in the Table 2. The AUROC results demonstrate a clear performance hierarchy across the 5 models, with the C-CKD Teacher achieving exceptional performance at 0.985 (95% CI: 0.975-0.993) when using both structured EHR and note data for both training and deployment. The knowledge distillation approach shows meaningful success, with the C-CKD Student model achieving 0.705 (95% CI: 0.668-0.743) and CKD model achieving 0.696 (0.656-0.733) using only structured EHR data at deployment time despite being trained on both structured EHR and note data, representing around 5% improvement over the baseline EHR Data Only model's 0.656 (95% CI: 0.616-0.692). The Contrastive model, which also trains on both data types but deploys on EHR only, achieves intermediate performance at 0.687 (95% CI: 0.649-0.726), outperforming the EHR Data Only baseline by 3.1% but falling short of the CKD approach. These results indicate that CKD loss effectively transfers information from the note-based teacher model to improve EHR-only predictions, while the contrastive loss provides more modest improvements.

**Table 1: demographic characteristics for late talk and control**

|  | Full Data Case | Full Data Control | Deployment Data Case | Deployment Data Control |
|---|---|---|---|---|
| **Number of Patients** | 1733 | 1733 | 347 | 347 |
| **Age Range (mths)** |  |  |  |  |
| 12~18 | 406 (23.4%) | 406 (23.4%) | 79 (22.8%) | 79 (22.8%) |
| 18~24 | 620 (35.8%) | 620 (35.8%) | 127 (36.6%) | 127 (36.6%) |
| 24~30 | 463 (26.7%) | 463 (26.7%) | 98 (28.2%) | 98 (28.2%) |
| **Sex** |  |  |  |  |
| Male | 1139 (65.7%) | 1139 (65.7%) | 236 (68.0%) | 236 (68.0%) |
| Female | 594 (34.3%) | 594 (34.3%) | 111 (32.0%) | 111 (32.0%) |

| Language | | | | |
|---|---|---|---|---|
| English | 1410 (81.4%) | 1410 (81.4%) | 292 (84.1%) | 292 (84.1%) |
| Spanish | 307 (17.7%) | 307 (17.7%) | 51 (14.7%) | 51 (14.7%) |
| Other | 16 (0.9%) | 16 (0.9%) | 4 (1.2%) | 4 (1.2%) |
| **Payer Type** | | | | |
| Public | 1050 (60.6%) | 1050 ((60.6%) | 213 (61.4%) | 213 (61.4%) |
| Private | 644 (37.2%) | 644 (37.2%) | 126 (36.3%) | 126 (36.3%) |
| Self-Pay/Special | 39 (2.3%) | 39 (2.3%) | 8 (2.3%) | 8 (2.3%) |
| **Race & Ethnicity** | | | | |
| Hispanic | 581 (33.5%) | 581 (33.5%) | 98 (28.2%) | 98 (28.2%) |
| NHBlack | 481 (27.8%) | 481 (27.8%) | 116 (33.4%) | 116 (33.4%) |
| NHWhite | 456 (26.3%) | 456 (26.3%) | 87 (25.1%) | 87 (25.1%) |

**Table 2: Model Performance Across Training and Deployment Modalities and Methods**

| Model | Training Modalities | Deployment Modalities | AUROC (95% CI) |
|---|---|---|---|
| C-CKD Teacher | Structured EHR + Notes | Structured EHR + Notes | 0.985 (0.975,0.993) |
| C-CKD Student | Structured EHR + Notes | Structured EHR | 0.705 (0.668,0.743) |
| Baseline: EHR Data Only | Structured EHR | Structured EHR | 0.656 (0.616,0.692) |
| Ablation 1: Contrastive | Structured EHR + Notes | Structured EHR | 0.687 (0.649,0.726) |
| Ablation 2: CKD | Structured EHR + Notes | Structured EHR | 0.696 (0.656,0.733) |

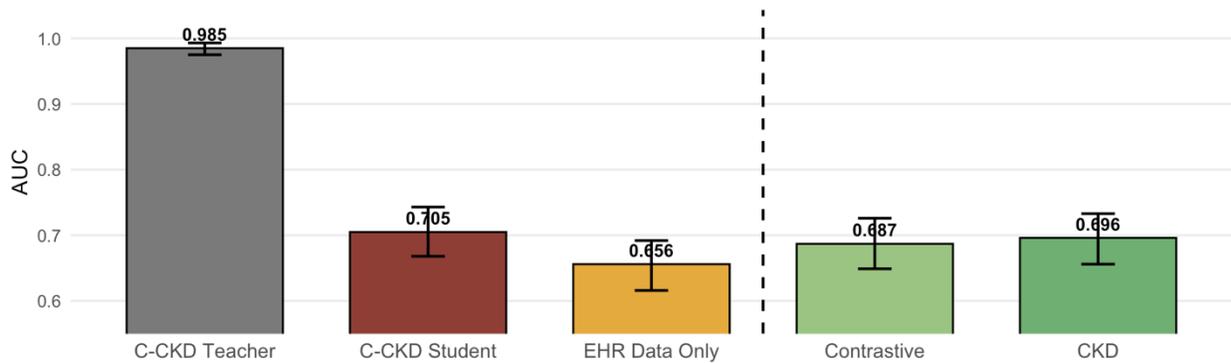

**Figure 2: AUROC comparison for C-CKD Teacher, C-CKD Student, EHR Only, Contrastive Model and CKD model.**

## Principal finding: impact on decision making

The performance metrics across varying alpha values for each model are shown in Figure 3. Coverage increases consistently for all models as probability threshold rises from 0.1 to 0.5, with the C-CKD Teacher achieving the highest coverage rates, followed by the C-CKD Student model, while EHR Data Only model showing the lowest coverage. Sensitivity demonstrates a clear trade-off relationship where higher probability thresholds lead to improved sensitivity, with C-CKD Teacher maintaining superior performance throughout, followed by C-CKD Student, and EHR Data Only showing the lowest sensitivity rates. Compared to sensitivity, specificity exhibits a different trend, while C-CKD Teacher again showing the best performance, C-CKD Student performs the best when probability threshold less or equal to 0.4. When probability threshold exceeds 0.4, the EHR Data Only model outperforms C-CKD Student slightly.

The Positive Predictive Value (PPV) and Negative Predictive Value (NPV) patterns are particularly informative for clinical interpretation. PPV remains relatively stable across probability thresholds for all models, with C-CKD Teacher maintaining the highest PPV, while other models cluster around 0.6 to 1.0. NPV shows a similar trend, with C-CKD Teacher consistently outperforming others, while other models cluster around 0.6 to 1.0. The accuracy metric reveals the overall trade-offs, where C-CKD Teacher demonstrates superior performance across all probability thresholds, maintaining accuracy near 1.0 even at higher alpha values, followed by C-CKD Student, and EHR Data Only showing the lowest accuracy. This comprehensive comparison suggests that the C-CKD Teacher model provides the most robust performance across different operational requirements, which aligns our expectation since it contains complete data modalities during training and deployment. The knowledge distillation approach (CKD) shows meaningful improvements over the baseline EHR only model in sensitivity, coverage, and accuracy, and the Contrastive only model offers competitive performance particularly in specificity, PPV and NPV metrics.

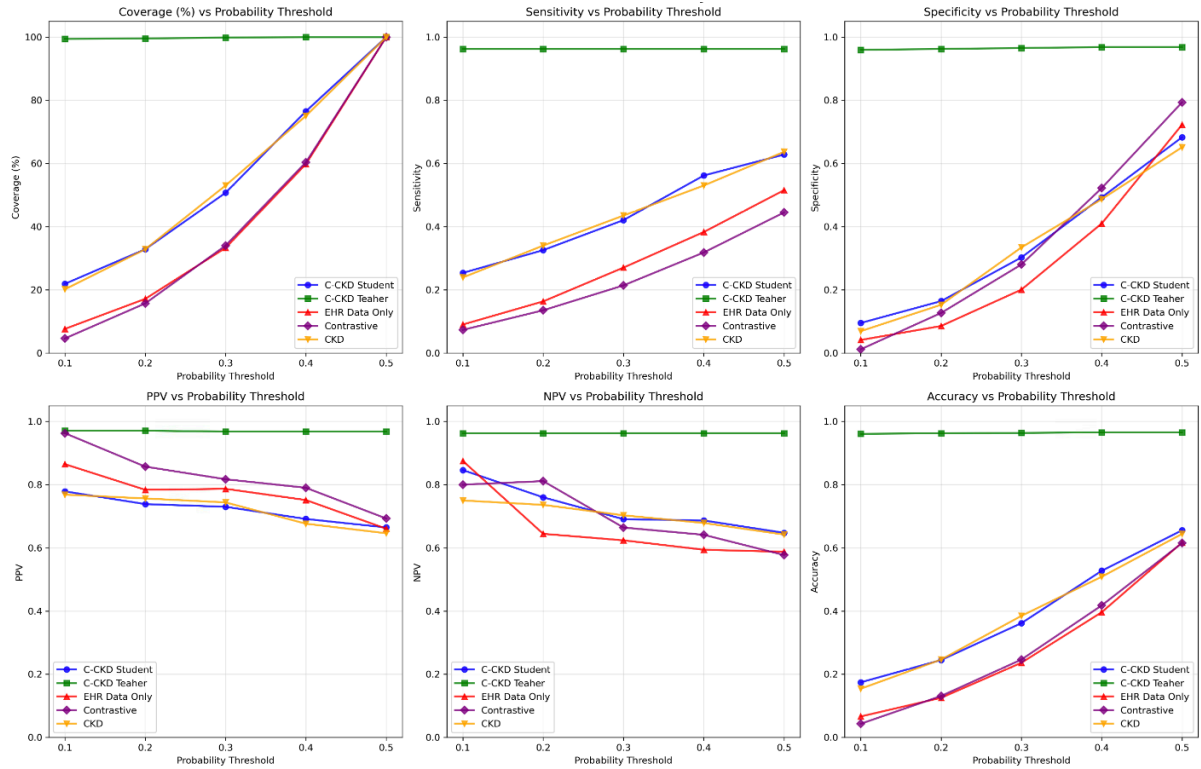

**Figure 3: Performance comparison for contrastive, knowledge distillation, and two baseline model (complete data modalities and structured EHR only)**

As shown in Table 2 and Figure 2, the AUROC results demonstrated similar trends across all configurations. The C-CKD Teacher—trained and evaluated with access to both structured EHR and clinical notes—achieved the highest discrimination performance (AUROC = 0.985, 95% CI: 0.975–0.993), reflecting near-perfect performance when all modalities are available. Among models operating on structured EHR data alone at deployment time, the C-CKD Student attained the strongest performance (AUROC = 0.705, 95% CI: 0.668–0.743), outperforming the baseline EHR Data Only (AUROC = 0.656, 95% CI: 0.616–0.692). The CKD model (AUROC = 0.696, 95% CI: 0.656–0.733) performed similarly to the C-CKD Student, with only a marginal 0.9% difference in AUROC. This pattern suggests that the primary source of improvement arises from the knowledge distillation loss, which transfers task-relevant predictive information from the note-based teacher model. The contrastive alignment loss component provided only modest incremental benefit by refining cross-modal representation similarity without substantially enhancing downstream discrimination performance.

## Discussion

### Principal findings

In this study, we developed a multimodal learning framework, called C-CKD, that leverages unstructured clinical notes during training while relying on structured EHR data alone for deployment. Our method improves structured-only modeling by (1) aligning structured and note representations through contrastive learning, and (2) distilling the pretrained teacher model's prediction patterns into the student model through contrastive knowledge distillation. This process enables the C-CKD Student model to learn richer, more predictive information from structured data than would be possible using structured features alone. As a result, our method achieves substantially improved predictive performance compared to conventional structured-only models. Specifically, the C-CKD Student model achieved an AUROC of 0.705, around 5% improvement over the baseline. While this improvement may seem modest, it is noteworthy given that our deployment setting assumes the absence of unstructured data at deployment time. In contrast, parallel contrastive learning or knowledge distillation frameworks typically operate under more favorable conditions (assuming both data modalities are available during training and testing), yet often report comparable or smaller improvement, typically below 5% [26,37]. Other studies under similar multimodal assumptions have demonstrated performance improvements within 1%, underscoring that even modest gains in this constrained deployment setting reflect meaningful enhancement in applicability [22]. Importantly, in the context of late-talking identification, a 5% increase can meaningfully improve early identification of late talkers, especially for children whose expressive-language concerns appear only in narrative notes and not in ICD codes. By capturing these otherwise overlooked cases, the structured-only model can reduce delays in evaluation and early intervention [38]. The results also highlight the importance of utilizing knowledge distillation. While contrastive learning improved model discrimination by aligning modality representations, it primarily focused on representation similarity rather than predictive alignment. It is understandable that not all note representations are useful for the prediction task. To filter out useful information, our knowledge distillation approach explicitly transferred task-specific information from the teacher model to the student model, leading to stronger performance under structured-only inference conditions.

Our approach addresses a critical challenge in EHR-based computable phenotyping—most multimodal models require access to all modalities, yet unstructured data like clinical notes are often unavailable at prediction time due to privacy restrictions, inconsistent documentation, or institutional barriers. Previous studies have proposed strategies for handling missing modalities or integrating multimodal data; however, these typically assume partial data availability at both training and inference [39]. In contrast, our framework targets the realistic scenario where all modalities are available during model training but only structured data are accessible at deployment/prediction time. By leveraging auxiliary unstructured data through contrastive learning and knowledge distillation, our method enhances model performance while preserving deploy ability. This work also holds broader implications for real-world EHR-based modeling. Clinical notes are often withheld from operational models due to privacy regulations and infrastructure constraints, and in other cases they are hard to collect, yet they capture contextual information critical to accurate phenotyping [23]. Our framework provides a way to indirectly benefit from unstructured text without requiring access at inference, thereby reducing the ethical and computational burden associated with direct text processing. Such strategies can facilitate the broader translation of multimodal research models into production environments.

## Limitations

Several limitations warrant consideration. First, our study focused on a single late-talking cohort, which may limit the generalizability of the findings to other phenotypes, age groups, or clinical settings. Replication in diverse populations and disease contexts is necessary to confirm the robustness of the proposed framework. Second, in practice, structured EHR data remain the most widely available data modality across institutions [23]. Therefore, our approach is designed to leverage additional modalities, such as unstructured clinical text, to enhance models that are ultimately deployed using structured data alone. This framework assumes that the teacher model trained on unstructured or other modality data provides superior or complementary information relative to the structured-only model. However, in cases where the teacher model performs comparably to or worse than the student model, the knowledge distillation process may yield limited or no performance improvement.

## Conclusion

In conclusion, our results demonstrate that our proposed framework enables computable phenotyping to benefit from data modalities available only during training, while remaining fully deployable using structured EHR data alone. By integrating contrastive learning for embedding alignment and knowledge distillation for task-specific knowledge transfer, the approach effectively bridges the gap between multimodal research models and real-world clinical implementation. Ultimately, this design not only improves phenotyping performance but also enhances model generalizability and in real-world settings where data completeness cannot be guaranteed.

# References

1. Pivovarov R, Perotte AJ, Grave E, Angiolillo J, Wiggins CH, Elhadad N. Learning probabilistic phenotypes from heterogeneous EHR data. J Biomed Inform 2015;58:156–165.
2. Wu J, Roy J, Stewart WF. Prediction Modeling Using EHR Data: Challenges, Strategies, and a Comparison of Machine Learning Approaches [Homepage on the Internet]. 2010; Available from: https://about.jstor.org/terms
3. Niu S, Ma J, Bai L, Wang Z, Guo L, Yang X. EHR-KnowGen: Knowledge-enhanced multimodal learning for disease diagnosis generation. Information Fusion 2024;102.
4. Kim MK, Rouphael C, McMichael J, Welch N, Dasarathy S. Challenges in and Opportunities for Electronic Health Record-Based Data Analysis and Interpretation. Gut Liver. 2024;18(2):201–208.
5. Si Y, Du J, Li Z, et al. Deep Representation Learning of Patient Data from Electronic Health Records (EHR): A Systematic Review.
6. Liu Z, Zhang J, Hou Y, Zhang X, Li G, Xiang Y. Machine Learning for Multimodal Electronic Health Records-based Research: Challenges and Perspectives.
7. Tayefi M, Ngo P, Chomutare T, et al. Challenges and opportunities beyond structured data in analysis of electronic health records. Wiley Interdiscip. Rev. Comput. Stat. 2021;13(6).
8. He T, Belouali A, Patricoski J, et al. Trends and opportunities in computable clinical phenotyping: A scoping review. J. Biomed. Inform. 2023;140.
9. Wang L, Olson JE, Bielinski SJ, et al. Impact of Diverse Data Sources on Computational Phenotyping. Front Genet 2020;11.
10. Pathak J, Kho AN, Denny JC. Electronic health records-driven phenotyping: challenges, recent advances, and perspectives [Homepage on the Internet]. Available from: http://phekb.org.
11. Pfaff ER, Crosskey M, Morton K, Krishnamurthy A. Clinical annotation research kit (CLARK): Computable phenotyping using machine learning. JMIR Med. Inform. 2020;8(1).
12. Patra BG, Sharma MM, Vekaria V, et al. Extracting social determinants of health from electronic health records using natural language processing: A systematic review. Journal of the American Medical Informatics Association. 2021;28(12):2716–2727.
13. Rosenbloom ST, Denny JC, Xu H, Lorenzi N, Stead WW, Johnson KB. Data from clinical notes: a perspective on the tension between structure and flexible documentation. Journal of the American Medical Informatics Association [homepage on the Internet] 2011;18(2):181–186. Available from: https://doi.org/10.1136/jamia.2010.007237
14. Wang G, Li C, Wang W, et al. Joint Embedding of Words and Labels for Text Classification [Homepage on the Internet]. In: Gurevych I, Miyao Y, editors. Proceedings of the 56th Annual Meeting of the Association for Computational Linguistics (Volume 1: Long Papers). Melbourne, Australia: Association for Computational Linguistics, 2018; p. 2321–2331.Available from: https://aclanthology.org/P18-1216/
15. Mullenbach J, Wiegreffe S, Duke J, Sun J, Eisenstein J. Explainable Prediction of Medical Codes from Clinical Text [Homepage on the Internet]. In: Walker M, Ji H, Stent A, editors. Proceedings of the 2018 Conference of the North American Chapter of the Association for Computational Linguistics: Human Language Technologies, Volume 1 (Long Papers). New Orleans, Louisiana: Association for Computational Linguistics, 2018; p. 1101–1111.Available from: https://aclanthology.org/N18-1100/
16. Garriga R, Buda TS, Guerreiro J, Omaña Iglesias J, Estella Aguerri I, Matić A. Combining clinical notes with structured electronic health records enhances the prediction of mental health crises. Cell Rep Med 2023;4(11).
17. Wang Z, Wu Z, Agarwal D, Sun J. MedCLIP: Contrastive Learning from Unpaired Medical Images and Text. 2022;Available from: http://arxiv.org/abs/2210.10163
18. Alsentzer E, Murphy JR, Boag W, et al. Publicly Available Clinical BERT Embeddings. 2019;Available from: http://arxiv.org/abs/1904.03323
19. Lee J, Yoon W, Kim S, et al. BioBERT: A pre-trained biomedical language representation model for biomedical text mining. Bioinformatics 2020;36(4):1234–1240.

20. Rasmy L, Xiang Y, Xie Z, Tao C, Zhi D. Med-BERT: pretrained contextualized embeddings on large-scale structured electronic health records for disease prediction. NPJ Digit Med 2021;4(1).
21. Zhang C, Cui Y, Han Z, Zhou JT, Fu H, Hu Q. Deep Partial Multi-View Learning. 2020;Available from: http://arxiv.org/abs/2011.06170
22. Xia M, Wilson J, Goldstein B, Henao R. Contrastive Learning for Clinical Outcome Prediction with Partial Data Sources.
23. Capurro D, Yetisgen M, Eaton E, Black R, Tarczy-Hornoch P. Availability of Structured and Unstructured Clinical Data for Comparative Effectiveness Research and Quality Improvement: A Multi-Site Assessment. eGEMs (Generating Evidence & Methods to improve patient outcomes) 2014;2(1):11.
24. He H, Hao W, Xi Y, Chen Y, Malin B, Ho JC. A Flexible Generative Model For Heterogeneous Tabular EHR With Missing Modality.
25. Radford A, Kim JW, Hallacy C, et al. Learning Transferable Visual Models From Natural Language Supervision. 2021;Available from: http://arxiv.org/abs/2103.00020
26. Chen T, Kornblith S, Norouzi M, Hinton G. A Simple Framework for Contrastive Learning of Visual Representations. 2020;Available from: http://arxiv.org/abs/2002.05709
27. Gou J, Yu B, Maybank SJ, Tao D. Knowledge Distillation: A Survey. Int J Comput Vis 2021;129(6):1789–1819.
28. Nouraey P, Ayatollahi MA, Moghadas M. Late language emergence a literature review. Sultan Qaboos Univ. Med. J. 2021;21(2):e182-190.
29. Shu J, Fannin DK, Dawson G, et al. Using Clinical Notes to Identify Late-Talking Children and Understand Differences in Diagnostic Timing. JAMIA Open 2025;
30. Hurst JH, Liu Y, Maxson PJ, Permar SR, Boulware LE, Goldstein BA. Development of an electronic health records datamart to support clinical and population health research. J Clin Transl Sci 2021;5(1).
31. Stolte A, Merli MG, Hurst JH, Liu Y, Wood CT, Goldstein BA. Using Electronic Health Records to understand the population of local children captured in a large health system in Durham County, NC, USA, and implications for population health research. Soc Sci Med [homepage on the Internet] 2022;296:114759. Available from: https://www.sciencedirect.com/science/article/pii/S0277953622000624
32. Wang H, Ma C, Zhang J, et al. Learnable Cross-modal Knowledge Distillation for Multi-modal Learning with Missing Modality. 2025;Available from: http://arxiv.org/abs/2310.01035
33. Zhu W, Zhou X, Zhu P, Wang Y, Hu Q. CKD: Contrastive Knowledge Distillation from A Sample-wise Perspective. 2025;Available from: http://arxiv.org/abs/2404.14109
34. Shlens J. Notes on Kullback-Leibler Divergence and Likelihood. 2014;Available from: http://arxiv.org/abs/1404.2000
35. Vest JR, Grannis SJ, Haut DP, Halverson PK, Menachemi N. Using structured and unstructured data to identify patients' need for services that address the social determinants of health. Int J Med Inform [homepage on the Internet] 2017;107:101–106. Available from: https://www.sciencedirect.com/science/article/pii/S1386505617302319
36. Huang J, Ling CX. Using AUC and Accuracy in Evaluating Learning Algorithms [Homepage on the Internet]. Available from: http://www.computer.org/publications/dlib
37. Wang F, Wang Y, Li D, et al. CL4CTR: A Contrastive Learning Framework for CTR Prediction. In: WSDM 2023 - Proceedings of the 16th ACM International Conference on Web Search and Data Mining. Association for Computing Machinery, Inc, 2023; p. 805–813.
38. Frelinger C, Gardner RM, Huffman LC, Whitgob EE, Feldman HM, Bannett Y. Detection of Speech-Language Delay in the Primary Care Setting: An Electronic Health Record Investigation. Journal of Developmental and Behavioral Pediatrics 2023;44(3):E196–E203.
39. Wu R, Wang H, Chen H-T, Carneiro G. Deep Multimodal Learning with Missing Modality: A Survey. 2024;Available from: http://arxiv.org/abs/2409.07825